\documentclass{article} 
\usepackage{main,times}

\usepackage{amsmath,amsfonts,bm}

\def\eqref#1{equation~\ref{#1}}

\def\1{\bm{1}}

\DeclareMathAlphabet{\mathsfit}{\encodingdefault}{\sfdefault}{m}{sl}
\SetMathAlphabet{\mathsfit}{bold}{\encodingdefault}{\sfdefault}{bx}{n}

\usepackage{graphicx}
\usepackage{hyperref}
\usepackage{url}
\usepackage{float}
\usepackage[most]{tcolorbox}
\usepackage{booktabs}
\usepackage{array}                 
\usepackage{makecell}              
\usepackage{xcolor}

\newcommand{\cellpass}[1]{\textbf{#1}}               
\newcommand{\cellnull}[1]{\textcolor{black!45}{#1}}
\newcommand{\cue}[1]{\textsc{#1}} 

\usepackage{amssymb}                       

\usepackage{fontawesome5}                      
\definecolor{sitePrompt}{HTML}{4E8F9C}
\definecolor{sitePixel}{HTML}{36618F}
\definecolor{siteContent}{HTML}{5E5A9B}
\definecolor{siteControl}{HTML}{8A918D}
\newlength{\biasiconw}
\newcommand{\biasrow}[3]{{\fontsize{6.6}{7.6}\selectfont%
  \makebox[\biasiconw][c]{\textcolor{#1}{#2}}#3}}
\newcommand{\floorrow}[2]{{\fontsize{6.6}{7.6}\selectfont%
  \makebox[\biasiconw][c]{\textcolor{siteControl}{#1}}\itshape\textcolor{black!55}{#2}}}
\newcommand{\alias}[1]{{\fontsize{6.0}{7.6}\selectfont\color{black!55}\,(#1)}}
\newlength{\marksep}
\newcommand{\dRho}{\rlap{\,$\scriptscriptstyle\blacktriangledown$}}
\newcommand{\uRho}{\rlap{\,$\scriptscriptstyle\blacktriangle$}}

\title{Do MLLM Judges Judge the Edit? Auditing Bias in Image Editing Evaluation with Verified Quality Preservation}

\author{Yuan Huang$^{1,2}$ \thanks{Contributed equally.}  \thanks{Visiting student at the Mohamed bin Zayed University of Artificial Intelligence.} 
\quad Zirui Song$^{2}$ \footnotemark[1]
\quad Xiuying Chen$^{2}$ \thanks{Corresponding author.}\\
Northeastern University$^{1}$ \\
Mohamed bin Zayed University of Artificial Intelligence$^{2}$\\
}

\usepackage{hyperref}

\iclrfinalcopy 
\begin{document}

\maketitle

\begin{abstract}
Multimodal large language models (MLLMs) are increasingly used as automated judges for instruction-based image editing and as reward signals for model training. However, systematically auditing whether these judges are influenced by cues irrelevant to editing quality is challenging because visual interventions may themselves alter the quality being evaluated. A judgment shift can therefore be attributed to bias only when the intervention is verified to preserve the underlying editing quality.
To address this challenge, we introduce EditJudgeBias, a counterfactual benchmark with verified quality preservation, comprising 1,196 real editing samples and 13 cues injected across four evaluation sites. We verify quality preservation for the requested edit using calibrated multimodal validators, controls, and human inspection. We then audit five MLLM judges along three complementary dimensions: invariance to quality-preserving cues, agreement with human judgments, and stability of pairwise preferences. Importantly, observed shifts are evaluated against each judge's own zero-dose and re-query noise floors rather than against zero.
Experiments show that quality-preserving cues move every judge beyond its own noise. Fabricated majority opinions increase ratings, irrelevant visual elements cause larger shifts than whole-image manipulations, and swapping candidate order reverses up to 60.9\% of pairwise decisions. Edit-region cues also tend to reduce human agreement. The three measures characterize judges differently, showing that robustness cannot be captured by a single metric.
\end{abstract}

\section{Introduction}
\label{sec:intro}

Recent advances in unified multimodal models have substantially improved instruction-based image editing~\citep{Bagel,show-o2,SenseNova-U1.5,Lance}. Evaluating these models requires judging instruction adherence, preservation of unrelated content, and visual quality. Because human evaluation is costly, benchmarks increasingly use multimodal models as automated judges ~\citep{RISE-Bench,KRIS-Bench}, while dedicated evaluators are trained on human preferences~\citep{wu2025editreward,ServeImage}. These judges are also used for output selection and reinforcement learning~\citep{luo2026editscore}, making their sensitivity to the evaluation procedure consequential. Do they measure the edit, or do quality-preserving cues move their decisions?

\begin{figure}[t]
    \centering
    \includegraphics[width=\linewidth]{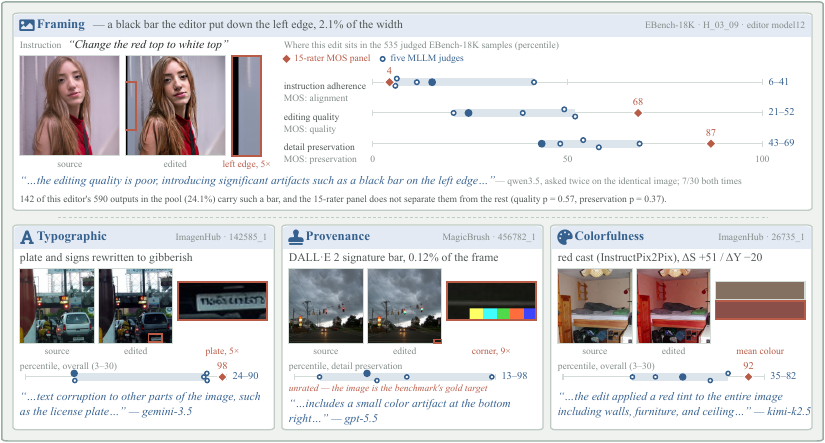}
    \caption{\textbf{Bias-relevant cues occur naturally in image-editing benchmarks.} Four real examples illustrate framing, typographic, provenance, and colorfulness cues. Human ratings may not penalize these cues, yet MLLM judges explicitly attend to them; EditJudgeBias tests these effects under controlled interventions}
    \label{fig:case1}
\end{figure}

Agreement with human judgments provides only a partial view of judge reliability~\citep{song2025beyond,xu2025socialmaze}. A judge may align well with humans on the original benchmark yet change its decision when the same edit is presented differently. Attributing an output to a particular model or claiming that other evaluators prefer it, for example, provides no new evidence about whether the instruction was followed. Visual cues are harder to test because the intervention itself may affect the edit: a region annotation can obscure an object, and a brightness adjustment can change an attribute requested by the instruction. Figure~\ref{fig:case1} illustrates why preservation has to be checked as part of the audit.

We introduce \textbf{EditJudgeBias}, a counterfactual benchmark for testing these sensitivities under controlled interventions. It contains 1,196 real editing triplets drawn from five public benchmarks and three independent content pools, with thirteen cues inserted at four points in the evaluation pipeline: the display protocol, judge prompt, image pixels, and edited content. Each sample is judged both with and without the cue. We collect three 1--10 rubric ratings in the scoring setting and preferences between two edits of the same request in the pairwise setting. Image-side cues are checked with calibrated validators and human annotations, and observed shifts are interpreted relative to each judge's \cue{sham} and re-query variability.

Quality-preserving cues systematically affect all five judges beyond their own response noise. In scoring, a fabricated majority opinion raises all three ratings across all judges, while a small sticker causes larger shifts than any whole-image manipulation. These effects also extend to human agreement: when a cue significantly changes agreement, it almost always reduces it, particularly for cues involving the edit region. In pairwise evaluation, simply swapping the display order reverses up to 60.9\% of decisions, far exceeding the 1.0 to 10.5\% reversal rate under repeated queries. Moreover, the judges behave differently across invariance, agreement, and stability, suggesting that no single robustness measure captures all failure modes.

\section{Related work}
\label{sec:related}

\subsection{Judges for instruction-based image editing}

Instruction-based image editing~\citep{brooks2023instructpix2pix} is commonly evaluated on benchmarks that pair edited images with human labels~\citep{zhang2023magicbrush,ku2024imagenhub,jiang2024genai,ma2024i2ebench,xu2025lmm4edit,ServeImage}. VIEScore helped establish rubric-prompted MLLM scoring for this setting~\citep{ku2024viescore}, an approach now used in leaderboards~\citep{liu2025step1x,ye2026imgedit}, agentic evaluators~\citep{wang2025unified}, and open judges~\citep{lee2024prometheus}. A parallel line of work trains editing-specific reward models against human preferences and uses them for online reinforcement learning~\citep{luo2026editscore,zhao2026trust,chen2026reasonedit,xu2026miescore,wang2025unifiedrm}. Other studies compare judges with human labels through factor-level agreement~\citep{liu2026human}, expert mean-opinion scores and intermediate-layer probing~\citep{gao2026evaluating,sun2026fine}, or dedicated preference suites~\citep{bai2026edit}.

Sensitivity to quality-preserving changes in evaluation has received much less attention. EditReward reports a position effect for one judge in an appendix~\citep{wu2025editreward}; GEditBench v2 randomizes presentation order without quantifying the resulting position bias~\citep{jiang2026geditbench}; and EdiVal-Agent catalogs failure modes of zero-shot judges without controlled interventions~\citep{chen2025edival}. They leave open whether presentation changes that preserve the underlying edit can move judge decisions beyond ordinary response variability. The issue also matters when judges are used as reward models, since optimization can amplify exploitable shortcuts~\citep{gao2023scaling}.

\subsection{Biases in MLLM-as-a-judge}
Text-only judges are known to exhibit position, verbosity, and self-preference biases \citep{zheng2023judging,wang2024large,saito2023verbosity,panickssery2024llm}. CALM systematically studies twelve such biases using robustness and consistency measures \citep{ye2025justice}. Recent work also shows that high consistency across repeated evaluations can coexist with substantial position bias \citep{norman2026reliability, xu2026inside}, motivating construct validity as a framework for evaluating judges \citep{chen2026judge}.

Multimodal judges face additional biases arising from visual inputs. In text-to-image (T2I) evaluation, FRAME studies eight visual manipulations and shows that they can inflate judge scores \citep{hwang2025fooling}. Other studies investigate adversarial perturbations \citep{wang2026adversarial}, preference bias, and prototypicality bias \citep{chen2026mj,roy2026prototypicality}. In visual question answering and image captioning, MM-JudgeBias evaluates nine types of perturbations across 26 MLLMs \citep{lee2026mm}, while other studies examine informativeness, perceptual, model-preference, and order biases \citep{zou2026vision,park2026mitigating,koyama2026mllm,ianaro2026order}. 
However, these audits focus on generated images, answers, or captions rather than instruction-based image editing. Evaluating an edit requires assessing whether the edited image faithfully follows the instruction given the source image, which introduces editing-specific factors such as the edit's location and extent. Cues tied to the edit region therefore remain largely unexplored in existing bias audits.

\subsection{Quality verification of injected cues}

Perturbation-based audits are interpretable only when the intervention leaves the evaluated construct unchanged; otherwise, a changed score may reflect a real degradation rather than evaluator bias. CheckList formalizes this as an invariance test~\citep{ribeiro2020beyond}. For LLM judges, \citet{chen2026judge} report both invariance under construct-preserving edits and sensitivity to construct-changing ones, while Perturbation CheckLists deliberately degrade outputs to test the latter~\citep{sai2021perturbation}. Preservation is handled less uniformly elsewhere: it may be assumed~\citep{vu2022layer}, checked on a small manual sample~\citep{wu2025rewordbench}, left unverified for rewritten variants~\citep{liu2025rm}, or avoided by collecting human ratings for both versions~\citep{huang2023robustness}.

Multimodal audits make different choices as well. FRAME does not verify that its manipulations are harmless and lists semantic-preserving perturbations as future work~\citep{hwang2025fooling}; MM-JudgeBias treats its composed perturbations as preservation-preserving~\citep{lee2026mm}; and \citet{agarwal2026image} verify 480 image--text items with three annotators each. Pixel or embedding similarity~\citep{wang2004image,zhang2018unreasonable,hessel2021clipscore} measures how much an image changes, but not whether the edited output still satisfies its instruction. Even near-invisible overlaid text can alter model behavior~\citep{cheng2024unveiling}. 
We evaluate preservation relative to the source image and instruction, using calibrated validators and human checks for every image-side cue.

\section{EditJudgeBias: Benchmark Design}
\label{sec:method}

\begin{figure}[t]
    \centering
    \includegraphics[width=\linewidth]{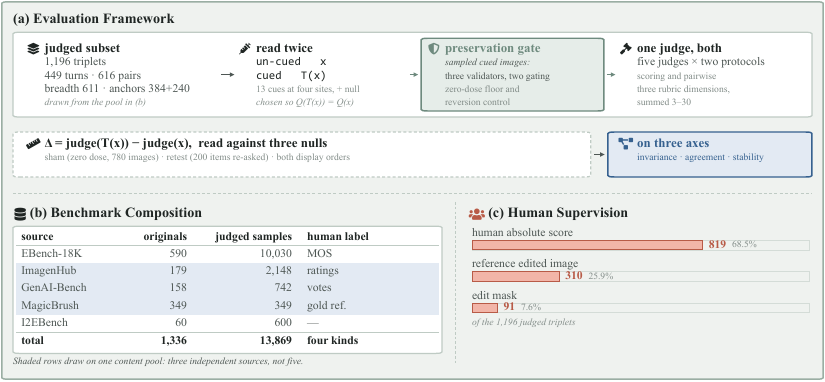}
    \caption{\textbf{Overview of EditJudgeBias.} (a) Evaluation framework for measuring judge invariance, agreement with human judgments, and stability under quality-preserving cues. (b) Composition of 1,196 editing triplets drawn from five public benchmarks spanning three independent content pools. (c) Human supervision used to assess agreement with human judgments and verify edit quality.}
    \label{fig:data_flow}
\end{figure}

Figure~\ref{fig:data_flow} summarizes the evaluation framework, benchmark composition, and human supervision used in EditJudgeBias.

\subsection{Problem Formulation}

Let $x=(I_{\mathrm{src}}, ins, I_{\mathrm{edit}})$ denote an editing sample comprising a source image, an editing instruction, and an edited image. Let $Q(x)$ denote its underlying quality across the three dimensions evaluated by the judges: instruction adherence, editing quality, and detail preservation. A judge $J$ estimates $Q$ under one of two protocols. In \emph{scoring}, $J(x)\in[3,30]$ is the sum of three 1--10 ratings, one per dimension. In \emph{pairwise} evaluation, $J(x_1,x_2)\in\{\mathrm{A},\mathrm{B},\mathrm{Tie}\}$ compares two edits of the same $(I_{\mathrm{src}}, ins)$, which is mainly dependent on $I_{\mathrm{edit}}^a$ and $I_{\mathrm{edit}}^b$.

We define a \emph{cue} as a transformation $T$ of the edited image or the evaluation procedure that preserves the underlying quality, so that $Q(T(x))=Q(x)$. A \emph{bias} arises when the judge's decision depends on this cue despite the preservation of $Q$. We distinguish the magnitude of this dependence from its direction: cue-induced changes are evaluated against judge-specific response-noise baselines, while signed-shift tests separately assess whether paired changes have a consistent direction (Section~\ref{sec:setup}).

For judge $j$, cue $c$, and rating dimension $d$, let $x_i^{(c)}$ and $x_i^{(0)}$ denote the cued and un-cued conditions, respectively. For the signed analysis, let $x_i^{(\mathrm{ref},c)}$ denote the cue-specific reference condition.
\begin{equation}
\textstyle
\begin{aligned}
I_{j,c,d} &= \textstyle \frac{10}{n^I_{j,c}} \sum_{i=1}^{n^I_{j,c}} \left| J_{j,d}(x_i^{(c)})-J_{j,d}(x_i^{(0)}) \right|,\\ 
\Delta_{j,c,d} &= \textstyle  \frac{1}{n^\Delta_{j,c}} \sum_{i=1}^{n^\Delta_{j,c}} \left[ J_{j,d}(x_i^{(c)}) - J_{j,d}(x_i^{(\mathrm{ref},c)}) \right].
\end{aligned}
\label{eq:scoring-effects}
\end{equation}
A judge that faithfully measures $Q$ should be invariant to $T$. In scoring, its ratings on all three dimensions should remain unchanged, implying $J(T(x))=J(x)$. In pairwise evaluation, $J(T(x_1),x_2)=J(x_1,x_2)$; if $T$ swaps the presentation order, $J(x_2,x_1)$ should return the corresponding mirrored preference. We assess \emph{invariance} by the mean absolute per-item change in each rating dimension, \emph{agreement} by the change in agreement with human judgments, and \emph{stability} by preference preservation under display-order swaps.

\subsection{Editing Samples}

We build EditJudgeBias from five published image-editing benchmarks with human supervision: I2EBench~\citep{ma2024i2ebench}, MagicBrush~\citep{zhang2023magicbrush}, ImagenHub~\citep{ku2024imagenhub}, GenAI-Bench~\citep{jiang2024genai}, and EBench-18K~\citep{xu2025lmm4edit}. Although these are five benchmarks, they span only three independent content pools because the rated ImagenHub images are drawn from the MagicBrush development set and most GenAI-Bench editing prompts come from the same pool. Their annotations provide the supervision needed for different parts of the audit, including per-dimension mean-opinion scores, absolute ratings, pairwise preferences, and gold edits (Appendix~\ref{app:protocol}).

The selected samples are organized around two complementary uses. A \emph{breadth} block of 611 samples, stratified by source and edit type, supports the invariance analysis across diverse edits. Two \emph{anchor} blocks instead preserve complete editing turns, keeping multiple editors' outputs for the same source image and instruction so that judge scores can be compared with human judgments. The anchor blocks contain 384 EBench-18K samples and 240 ImagenHub samples.

\begin{figure}[t]
    \centering
    \includegraphics[width=\linewidth]{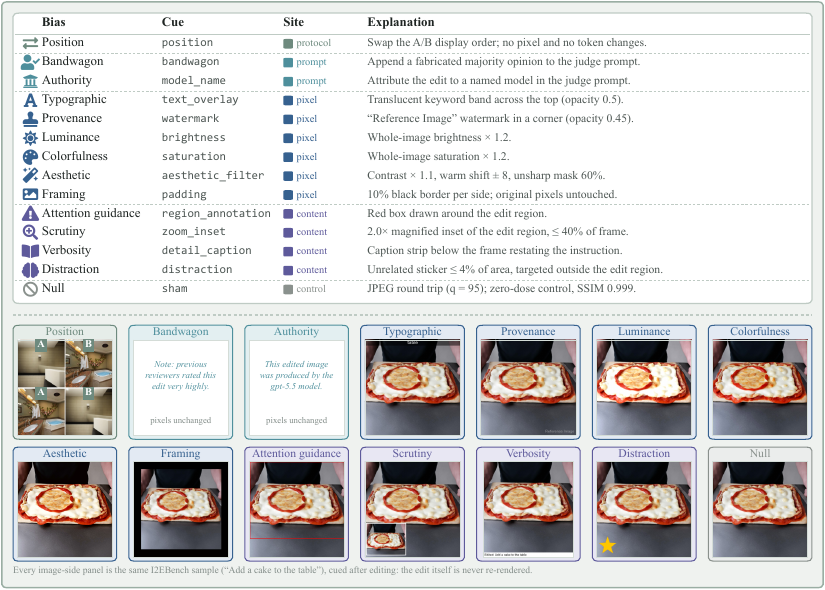}
    \caption{\textbf{Overview of the thirteen bias cues and the zero-dose control.} Each row identifies a cue, its injection site in the judging pipeline, and its intervention and dose. All image-side cues are illustrated using the same edited I2EBench example. \cue{position} is illustrated by the same judged pair under both display orders; it changes neither pixels nor prompt text.}
    \label{fig:showcase}
\end{figure}

\subsection{Bias Cues}
\label{sec:cues}

We introduce 13 controlled cues into the selected editing samples, grouped by where they enter the judging pipeline: protocol, prompt, pixel, or content as shown in Figure~\ref{fig:showcase}. This grouping identifies each intervention's entry point and helps localize where an observed effect may arise. Protocol and prompt cues introduce no visual changes and extend biases studied in text-only evaluation \citep{ye2025justice,panickssery2024llm}. Pixel cues modify image pixels through photometric, framing, or localized overlay operations~\citep{hendrycks2019imagenetc,cheng2024unveiling}. Content cues are defined relative to the edit region and are specific to image-editing evaluation. Exact operators and doses are provided in Appendix~\ref{app:protocol}.

For content cues, we use the benchmark-provided source mask when available; otherwise, we estimate the edit region as the largest connected component of the thresholded difference between the source and edited images.
We apply pixel and content cues deterministically to the edited image, without re-rendering the underlying edit. We also include \cue{sham}, a zero-dose control that provides the judge-specific sham floor used to contextualize cue-induced shifts. Our pixel-level cues partially overlap with those studied in FRAME and MM-JudgeBias \citep{hwang2025fooling,lee2026mm}.

\subsection{Quality-Preservation Verification}
\label{sec:preservation}

Attributing a cue-induced judgment shift to bias requires establishing that the cue preserves underlying editing quality, i.e., $Q(T(x))=Q(x)$. Without this condition, the shift could reflect a genuine change in quality rather than bias. We assess preservation using three complementary sources of evidence: image similarity, calibrated MLLM validators, and human checks.

\paragraph{Similarity.}
For every cued image, we report whole-image SSIM \citep{wang2004image} against its un-cued counterpart. For content-site cues, SSIM also captures the added overlay itself and therefore provides only a conservative measure of preservation. We use SSIM as a descriptive measure rather than a filtering criterion.

\paragraph{Calibrated MLLM validators.}

We use MLLM validators to determine whether a cue changes the underlying edit. To check that they distinguish preserved edits from genuine quality changes, we calibrate them in both directions. The zero-dose \cue{sham} condition measures each validator's false-positive rate, while positive controls measure its sensitivity to known quality changes. For the positive controls, we revert the edit region toward the source image at three strengths and separately blur the edit region. These interventions provide known violations of preservation without requiring human ground truth.

We evaluate three validators on 110 randomly sampled images per cue and 780 \cue{sham} images. For the filtered analysis, we retain only images that pass the two validators with the lower \cue{sham} flag rates. The third validator, from a model family absent from the judge panel, is designed to provide an independent cross-check.

\paragraph{Human checks.}
We also assess preservation for all ten image-side cues through human annotation, using 30 examples per cue. The first set contains 150 examples covering five pixel-site cues. The second contains 180 examples covering four content-site cues, one pixel-site cue, and 30 \cue{sham} controls; cue identities remain hidden until annotation is complete. Annotators label whether each cue changes the underlying edit as No, Slightly, or Yes. Under the pre-defined criterion, only Yes counts as a preservation failure. Two annotators work independently, with the second annotating a 60-item subset of the blinded set. 
Appendix~\ref{app:validator} reports the validator pass rates, calibration against positive controls, human-check results, and inter-annotator agreement.

\section{Experimental setup}
\label{sec:setup}

\subsection{Evaluation Metrics}

\emph{Invariance.}
For each judge and cue, we measure the mean absolute paired change in each 1--10 rating dimension between the un-cued and cued conditions, reported as 10 times the mean absolute paired change. We also report the signed paired mean shift $\Delta$ to show the direction of the effect, assess significance with the Wilcoxon signed-rank test, and report bootstrap confidence intervals. To distinguish cue effects from intrinsic variability, we compare the absolute changes against two judge-specific baselines: the change under \cue{sham} and retest noise, estimated by re-evaluating 200 un-cued samples.

For $b\in\{\mathrm{sham},\mathrm{retest}\}$, let $\widehat I_{j,d}^{(b,r)}$ denote the mean absolute paired change (multiplied by 10) in bootstrap resample $r$. We define each null floor as
\begin{equation}
\textstyle
F_{j,d}^{(b)} = \textstyle Q_{0.20}\!\left( \left\{\widehat I_{j,d}^{(b,r)}\right\}_{r=1}^{2000} \right), \qquad b\in\{\mathrm{sham},\mathrm{retest}\},
\label{eq:noise-floors}
\end{equation}
where $Q_p$ denotes the empirical $p$-quantile. Let $I^{\mathrm{lo}}_{j,c,d}$ denote the lower end of the corresponding 95\% bootstrap interval for the cue-induced absolute change. A cue--dimension cell clears the noise criterion when
\begin{equation}
\textstyle
I^{\mathrm{lo}}_{j,c,d} > \max \left\{ F^{\mathrm{sham}}_{j,d}, F^{\mathrm{retest}}_{j,d} \right\}.
\label{eq:noise-criterion}
\end{equation}
\emph{Agreement.}
We measure agreement with human judgments using Spearman's correlation between judge scores and human labels, and denote the cue-induced change by $\Delta\rho$. For pairwise evaluation, we measure accuracy against human preferences and test changes with McNemar's test. We cluster bootstrap intervals by editing turn to account for dependence among outputs from the same source and instruction.

For anchor benchmark $a$ and rating dimension $d$, let $H_{d,a}$ denote the human labels and let $X_{c,a}$ and $X^{\mathrm{ref}}_{c,a}$ denote the cued and paired reference arms. The cue-induced agreement change is
\begin{equation}
\textstyle
\Delta\rho_{j,c,d,a} = \rho_{\mathrm S}\!\left(J_{j,d}(X_{c,a}),H_{d,a}\right) - \rho_{\mathrm S}\!\left(J_{j,d}(X^{\mathrm{ref}}_{c,a}),H_{d,a}\right).
\label{eq:agreement}
\end{equation}
\emph{Stability.}
Following CALM \citep{ye2025justice}, we measure pairwise stability using the robustness rate, which captures whether preferences are preserved after swapping the A/B presentation order, and the consistency rate, which measures agreement across repeated evaluations. We also report the slot-preference rate to quantify systematic preference for a presentation position.
Let $m(\mathrm{A})=\mathrm{B}$, $m(\mathrm{B})=\mathrm{A}$, and $m(\mathrm{Tie})=\mathrm{Tie}$. Pairwise stability under an A/B swap is
\begin{equation}
\textstyle
S_j = \textstyle \frac{1}{N} \sum_{i=1}^{N} \mathbf{1} \left[ J_j(x_{i,2},x_{i,1}) = m\!\left(J_j(x_{i,1},x_{i,2})\right) \right].
\label{eq:stability}
\end{equation}
We apply Benjamini--Hochberg correction within each pre-defined family of comparisons. We exclude the \cue{sham} control from the invariance family and treat comparisons for the replication judge in Section~\ref{sec:judges} as a separate family.

\subsection{Judges and Evaluation Setup}
\label{sec:judges}
We audit five judges in three groups: two closed frontier models, gpt-5.5 and gemini-3.5-flash; two models from open-weight families, qwen3.5-plus and kimi-k2.5; and one task-specific judge, VIEScore \citep{ku2024viescore}, whose rubric we run on GPT-4o. We request temperature 0 whenever the parameter is configurable; for gpt-5.5, the provider fixes this setting. Retest noise is estimated separately for each judge. 
We also evaluate a sixth, open-weight judge, qwen3-vl-32b-instruct, on the same samples and cue conditions as a pre-declared replication. We analyze its results as a separate family of comparisons and do not pool them with those of the five primary judges. Appendix~\ref{app:protocol} lists the model strings and temperature settings.

\section{Results and discussion}
\label{sec:res}

\begin{table}[t]
\centering
\fontsize{7.5}{8.8}\selectfont
\setlength{\tabcolsep}{0pt}
    \caption{\textbf{Invariance and human agreement across cues and judges.}
    Cells report 10 times the mean absolute per-item change in each 1--10 rating dimension
    (IA: instruction adherence; EQ: editing quality; DP: detail preservation).
    Bold indicates cells whose 95\% bootstrap lower bound exceeds both the
    \cue{sham} and retest floors; plain cells exceed one floor and gray cells
    exceed neither.
    Superscripts indicate significant changes in agreement with human ratings
    ($\blacktriangledown$ decrease; $\blacktriangle$ increase).
}
\label{tab:main}
\begin{tabular}{@{}>{\raggedright\arraybackslash}p{96pt}@{\hspace{2pt}}*{3}{r@{\hspace{\marksep}}}@{\hspace{3.0pt}}*{3}{r@{\hspace{\marksep}}}@{\hspace{3.0pt}}*{3}{r@{\hspace{\marksep}}}@{\hspace{3.0pt}}*{3}{r@{\hspace{\marksep}}}@{\hspace{3.0pt}}*{3}{r@{\hspace{\marksep}}}}
\toprule
 & \multicolumn{3}{c}{gpt-5.5} & \multicolumn{3}{c}{gemini-3.5} & \multicolumn{3}{c}{kimi-k2.5} & \multicolumn{3}{c}{qwen3.5} & \multicolumn{3}{c}{VIEScore} \\
\cmidrule(lr){2-4}\cmidrule(lr){5-7}\cmidrule(lr){8-10}\cmidrule(lr){11-13}\cmidrule(lr){14-16}
Bias\,\textcolor{black!55}{\fontsize{6.0}{7.6}\selectfont(short name)} & IA & EQ & DP & IA & EQ & DP & IA & EQ & DP & IA & EQ & DP & IA & EQ & DP \\
\midrule
\biasrow{sitePrompt}{\faUserCheck}{Bandwagon} & 8.2 & \cellpass{11.5} & 9.0\dRho & \cellpass{18.5}\dRho & \cellpass{18.1} & \cellpass{18.5} & \cellpass{5.2} & \cellpass{6.0} & \cellpass{4.7} & \cellpass{3.5} & \cellpass{6.0} & \cellpass{5.7} & \cellpass{7.2} & \cellpass{7.3}\dRho & \cellpass{5.0}\dRho \\
\biasrow{sitePrompt}{\faUniversity}{Authority}\alias{model name} & \cellnull{6.4} & \cellnull{7.9} & \cellnull{6.6} & 17.8 & 15.1 & 17.3 & 2.9 & 2.9 & 3.3 & 3.2 & \cellpass{4.7} & 4.5 & 4.4 & 4.9 & 4.2 \\
\addlinespace[3pt]
\biasrow{sitePixel}{\faPalette}{Colorfulness}\alias{saturation} & \cellnull{5.8} & \cellnull{7.3} & \cellnull{6.0} & \cellnull{8.2}\dRho & \cellnull{7.5} & \cellnull{7.6} & 3.0 & 2.7 & 3.2 & \cellnull{2.4} & 3.9 & 4.3 & 4.2 & 4.7 & 3.8 \\
\biasrow{sitePixel}{\faMagic}{Aesthetic}\alias{aesthetic filter} & \cellnull{6.6}\dRho & 8.5\dRho & 7.8 & \cellnull{12.0} & \cellnull{10.8} & \cellnull{12.2} & \cellpass{3.8} & \cellpass{3.8} & \cellpass{4.6} & \cellpass{4.1} & \cellpass{5.3} & \cellpass{5.4} & \cellpass{6.1} & \cellpass{6.0} & \cellpass{5.3} \\
\biasrow{sitePixel}{\faStamp}{Provenance}\alias{watermark} & \cellnull{6.8} & \cellnull{7.9}\dRho & \cellnull{7.4}\dRho & \cellnull{7.7}\dRho & \cellnull{8.0} & \cellnull{7.6} & 3.1 & \cellpass{3.3} & \cellpass{3.6} & \cellpass{4.6} & \cellpass{6.9} & \cellpass{6.2} & \cellpass{7.1} & \cellpass{8.2}\dRho & \cellpass{6.2} \\
\biasrow{sitePixel}{\faSun[solid]}{Luminance}\alias{brightness} & 6.9 & 8.5 & \cellnull{7.2} & \cellnull{7.7} & \cellnull{7.5} & \cellnull{7.8} & \cellpass{3.7} & \cellpass{4.1} & \cellpass{5.1} & \cellpass{3.9} & \cellpass{5.7} & \cellpass{6.1} & 5.1 & \cellpass{6.0} & \cellpass{5.0} \\
\biasrow{sitePixel}{\faImage[solid]}{Framing}\alias{padding} & 7.3 & 9.5\dRho & \cellpass{13.5}\dRho & \cellnull{8.0} & \cellnull{8.1}\dRho & \cellnull{9.3} & \cellpass{4.6} & \cellpass{4.3} & \cellpass{7.1}\dRho & \cellpass{5.2} & \cellpass{7.8} & \cellpass{9.2} & \cellpass{7.8} & \cellpass{7.7} & \cellpass{6.7} \\
\biasrow{sitePixel}{\faFont}{Typographic}\alias{text overlay} & 9.8 & \cellpass{12.7} & \cellpass{11.4} & \cellnull{8.3} & \cellnull{9.5} & \cellnull{9.8} & \cellpass{7.1} & \cellpass{6.6} & \cellpass{5.8} & \cellpass{7.4} & \cellpass{10.8} & \cellpass{8.2} & \cellpass{8.9} & \cellpass{9.5}\dRho & \cellpass{6.5}\dRho \\
\addlinespace[3pt]
\biasrow{siteContent}{\faBookOpen}{Verbosity}\alias{caption} & \cellnull{6.3} & \cellnull{7.6} & \cellnull{7.4} & 17.8\dRho & 15.1 & 17.5 & \cellpass{4.1} & \cellpass{4.0} & \cellpass{4.3} & \cellpass{5.9}\uRho & \cellpass{7.7}\dRho & \cellpass{6.4} & \cellpass{9.3} & \cellpass{8.7} & \cellpass{6.6}\dRho \\
\biasrow{siteContent}{\faExclamationTriangle}{Attention guidance}\alias{box} & 9.4 & \cellpass{13.6}\dRho & \cellpass{13.6}\dRho & 14.9\dRho & \cellpass{16.5}\dRho & \cellnull{15.1}\dRho & \cellpass{4.6} & \cellpass{6.2} & \cellpass{6.4} & \cellpass{6.5} & \cellpass{9.0} & \cellpass{8.6} & \cellpass{11.7}\dRho & \cellpass{10.9} & \cellpass{7.5}\dRho \\
\biasrow{siteContent}{\faSearchPlus}{Scrutiny}\alias{inset}& \cellpass{11.4} & \cellpass{16.2}\dRho & \cellpass{18.1}\dRho & 14.9\dRho & \cellpass{21.7}\dRho & \cellpass{21.5}\dRho & \cellpass{8.3}\dRho & \cellpass{9.2}\dRho & \cellpass{10.1}\dRho & \cellpass{10.2} & \cellpass{13.7} & \cellpass{12.5}\dRho & \cellpass{12.2}\dRho & \cellpass{10.8}\dRho & \cellpass{8.8}\dRho \\
\biasrow{siteContent}{\faBrain}{Distraction}\alias{sticker} & \cellpass{10.8} & \cellpass{13.5} & \cellpass{19.5}\dRho & 14.8\dRho & 15.5\dRho & 17.6\dRho & \cellpass{10.5} & \cellpass{9.1} & \cellpass{8.9}\dRho & \cellpass{11.3} & \cellpass{13.7} & \cellpass{16.4} & \cellpass{10.3} & \cellpass{10.8}\dRho & \cellpass{8.1} \\
\midrule
\floorrow{\faBan}{sham floor} & \itshape 6.2 & \itshape 7.7 & \itshape 6.8 & \itshape 17.5 & \itshape 15.2 & \itshape 16.9 & \itshape 2.7 & \itshape 2.7 & \itshape 3.1 & \itshape 2.4 & \itshape 3.5 & \itshape 3.7 & \itshape 4.4 & \itshape 4.9 & \itshape 3.7 \\
\floorrow{\faRedo}{retest floor} & \itshape 9.2 & \itshape 10.3 & \itshape 9.1 & \itshape 13.7 & \itshape 13.0 & \itshape 15.1 & \itshape 2.0 & \itshape 1.8 & \itshape 2.0 & \itshape 1.9 & \itshape 2.5 & \itshape 3.4 & \itshape 2.9 & \itshape 3.8 & \itshape 2.9 \\
\bottomrule
\end{tabular}
\end{table}

\subsection{Invariance under Quality-Preserving Cues}
\label{sec:res-invariance}

\paragraph{Scoring.}
Quality-preserving cues can shift ratings beyond judge-specific response noise. Excluding \cue{inset}$^{\ddagger}$, half of the instruction-adherence cells in Table~\ref{tab:main} clear both the
\cue{sham} and retest floors; the corresponding counts are 34 of 55 for editing quality and 32 of 55 for detail preservation. The effects are not uniform across cues: \cue{bandwagon} clears both floors for most judge--dimension combinations, whereas \cue{saturation} never does. Despite covering at most 4\% of the frame, the sticker produces larger rating changes than any pixel-site cue in every judge--dimension cell.

The noise-floor criterion and the signed-shift test capture different properties of the response. A cell that does not clear both noise floors can still exhibit a significant signed shift: the former asks whether the magnitude of change exceeds response variability, whereas the latter tests whether paired changes show a consistent direction. A gray cell in Table~\ref{tab:main} should therefore not be read as evidence of no directional effect.

\paragraph{Pairwise.}
Several cues also shift pairwise preferences, showing that cue sensitivity is not specific to scalar scoring. Figure~\ref{fig:panels}a reports cue-induced changes in pairwise win rates. \cue{bandwagon} increases the cued candidate's win rate under every judge, while the sticker and text overlay tend to disadvantage. By contrast, saturation, the aesthetic filter, and the watermark have little effect on pairwise preferences. The effect of a quality-preserving cue therefore depends not only on the judge and cue, but on whether evaluation is elicited as an absolute score or a comparative preference.

\subsection{Agreement with Human Judgments}
\label{sec:res-agreement}

Table~\ref{tab:main} marks significant cue-induced changes in judge–human agreement, which are overwhelmingly negative.
Excluding the borderline INSET, 33 of the 34 significant changes are decreases. These decreases occur in 16 of 45 content-site comparisons, compared with 13 of 90 pixel-site and 4 of 30 prompt-site comparisons. This concentration among edit-region cues is descriptive rather than a formal comparison across sites, because the pixel and content groups each include comparisons against both SHAM and the original uncued images.

The pairwise results provide a more direct example. When \cue{bandwagon} assigns a fabricated majority to the candidate rejected by humans, agreement falls for every judge; reversing the fabricated majority instead increases agreement for all five. The same judge can appear more or less aligned with human preferences depending on a quality-irrelevant change in the evaluation context, illustrating why agreement alone is insufficient as a robustness measure.

\subsection{Pairwise Stability under Display-Order Swaps}
\label{sec:res-stability}

\begin{figure}[t] 
\centering 
\includegraphics[width=\linewidth]{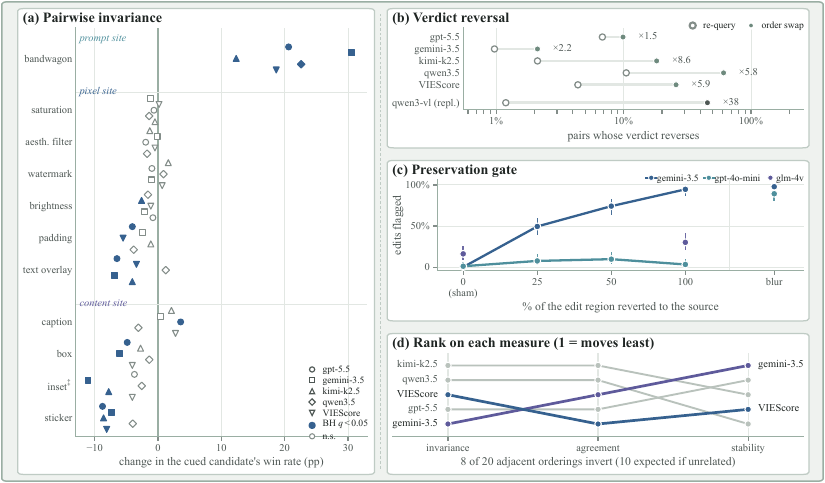} 
    \caption{\textbf{Main results of the EditJudgeBias audit.} (a) Pairwise preference shifts induced by quality-preserving cues. (b) Verdict reversal rates under display-order swaps, compared with identical re-queries. (c) Calibration of the preservation validators using controlled edit degradation and the zero-dose control. (d) Comparison of judges across invariance, agreement, and stability, showing that judge rankings differ across the three robustness measures.}
\label{fig:panels} 
\end{figure}

Figure~\ref{fig:panels}b compares verdict-reversal rates under display-order swaps and identical re-queries, showing substantially higher rates under order swaps.
This separation holds across the primary judge panel and is especially pronounced for qwen3.5, for which 60.9\% of order-swapped pairs reverse compared with 10.5\% under identical re-query. 
The replication judge shows the same qualitative separation between re-query variability and sensitivity to display order.

The two-order rule used by the MT-Bench~\citep{zheng2023judging} filters rather than resolves this instability: a winner is retained only when both presentation orders agree. This removes order-sensitive pairs at the cost of coverage, most sharply for qwen3.5. A lenient variant additionally retains the non-Tie choice when only one order returns Tie; pairs whose two orders select opposite winners remain unresolved. The Appendices~\ref{app:controls} and~\ref{app:replication} report the reconciliation analysis and the replication results, respectively.

\subsection{Preservation Evidence}
\label{sec:res-preservation}

Our preservation checks support treating most image-side cues as quality-preserving at the cue level. Figure~\ref{fig:panels}c shows the validators' responses to controlled degradations. Among the ten image-side cues, only \cue{inset} shows a significant departure from paired \cue{sham} images. In the sham-controlled human annotation set, none of the five evaluated cues differs significantly from \cue{sham} under the prespecified criterion.

The evidence for \cue{inset} is less clear. Two validators flag it more often than \cue{sham}, and a more inclusive post-hoc human criterion also identifies a difference. These signals do not justify treating the cue as uniformly quality-preserving, but neither do they overturn the preservation evidence for the remaining image-side interventions. We therefore treat \cue{inset} as borderline, report its results separately, and exclude it from aggregate counts. This separation prevents the ambiguous cue from affecting the main aggregate conclusions while retaining its individual results for transparency.

\subsection{Robustness across Measures}
\label{sec:res-axes}

Figure~\ref{fig:panels}d shows that the three robustness measures induce different judge orderings. kimi-k2.5 ranks first on invariance and agreement but third on stability, whereas gemini-3.5 moves from fifth on invariance to first on stability. Across adjacent measure pairs, 8 of 20 judge orderings reverse, compared with 10 under the unrelated-measure reference. These differences indicate that invariance, agreement, and stability capture distinct aspects of judge behavior rather than interchangeable estimates of a single robustness quantity. A single aggregate score would therefore obscure which failure mode is responsible for a judge's apparent weakness. Appendix~\ref{app:fill} reports the exact rankings and alternative summaries.

\section{Conclusion}
\label{sec:conc}

We introduced \textbf{EditJudgeBias} to test how image-editing judges respond to controlled cues that are designed to preserve the underlying edit. Across the five primary judges, such cues can shift both ratings and pairwise preferences beyond ordinary response variability. Human agreement does not fully predict this behavior: display-order swaps can reverse preferences, and the same judges rank differently on invariance, agreement, and stability.
More broadly, our findings suggest that judge reliability should not be defined solely by whether a judge agrees with humans, but also by whether its decisions remain invariant when the underlying quality being evaluated is unchanged. This offers an insight for MLLM-based evaluation: reliable judging requires measuring the intended construct rather than responding to incidental features of how that construct is presented.

\clearpage
\subsection*{AI Disclosure}

Generative AI tools assisted with drafting and language editing, figure preparation, and software tasks including code review and repository maintenance. The research questions, literature selection, experimental design and deployment, data analysis, and interpretation were carried out by the authors.

\subsection*{Ethics Statement}

EditJudgeBias uses existing published image-editing benchmarks and controlled interventions intended to audit automated evaluators. Image-side cues are checked for preservation, and human annotation is used only for this purpose. Some prompt-side cues deliberately present fabricated social information, such as a claimed majority preference, to test susceptibility to information irrelevant to editing quality. These statements are confined to the evaluation protocol and are not treated as factual benchmark annotations.

\subsection*{Reproducibility Statement}

We provide an open-source repository at \href{https://github.com/yuan4629/EditJudgeBias}{reproducibility repository}. This release contains code, configurations, prompts, and scripts, but excludes benchmark images, human annotation records, and experimental judge/validator responses. The author-generated I2EBench outputs used in our experiments are also not included. The release therefore does not by itself permit exact reconstruction of all evaluated inputs or reported results.

Several judges are accessed through external APIs, so repeated identical calls may produce different outputs. Our design explicitly accounts for this variability using judge-specific re-query baselines, \cue{sham} controls, paired inference, confidence intervals, multiple-testing correction, and a separately analyzed replication arm. Reproducibility therefore concerns the statistical findings and qualitative conclusions under expected API variability rather than exact reproduction of every model response.

\bibliography{main}
\bibliographystyle{main}

\clearpage
\appendix

\section{Limitations}

Our study has three limitations. First, EditJudgeBias evaluates a finite set of controlled cues at specified intervention strengths; interactions among co-occurring cues and responses across a broader range of strengths remain to be characterized. Second, the results reflect the evaluated judge versions and configurations. As models and API services evolve, effect magnitudes and relative judge rankings may change, motivating periodic re-evaluation with the same audit protocol. Third, our experiments focus on diagnosing cue sensitivity, while mitigation is explored only through limited aggregation analyses and a preliminary prompting pilot. Developing and systematically evaluating mitigation strategies under the main experimental protocol remains future work. 
Our preservation checks focus on the requested edit; rubric-aligned human ratings of both versions are needed to distinguish evaluator bias from legitimate visual-quality or detail-preservation penalties.

\section{Benchmark and Evaluation Protocol}
\label{app:protocol}

\subsection{Editing samples and pools}
\label{app:data}

The judged subset contains 1,196 samples drawn from five published editing benchmarks. These benchmarks represent three independent content pools rather than five: the rated ImagenHub images are a subset of the MagicBrush development set, and most GenAI-Bench editing prompts come from the same pool. Figure~\ref{fig:data_flow}b marks the benchmarks that share underlying content.

The subset is divided into one \emph{breadth} block and two \emph{anchor} blocks. The breadth block contains 611 samples stratified by source and edit type and is used for the invariance analysis. The anchor blocks retain complete editing turns, keeping multiple editors' outputs for each source--instruction pair together. EBench-18K contributes 48 turns with 8 of the block's 17 editors, for 384 samples; ImagenHub contributes 30 turns with 8 editors, for 240 samples. These blocks provide the human labels used in the agreement analysis and are also the blocks on which we re-score editor leaderboards. Across these blocks there are 1,235 sample assignments; 39 triplets occur in both the breadth and anchor blocks, yielding 1,196 unique triplets overall. The pairwise protocol uses 616 pairs of edits for the same request, drawn from 1,086 edited images; 438 pairs have a decisive human preference.

\subsection{Cues, operators and doses}
\label{app:cues}

All image-side cues are deterministic PIL operations applied after the edit has been produced. For the global pixel-site cues, \cue{saturation} and \cue{brightness} scale the corresponding enhancement by 1.2. The \cue{aesthetic filter} increases contrast by 1.1, shifts the red and blue channels by $+8$ and $-8$ points for a warm cast, and applies an unsharp mask at 60\% with radius 1/256 of the frame. \cue{padding} adds a black border equal to 10\% of each image dimension on every side. \cue{watermark} places ``Reference Image'' in the bottom-right corner at 45\% opacity, using a font size of 4\% of the short edge and a 3\% inset. \cue{text overlay} places a salient word from the instruction in a semi-transparent band along the top edge at 50\% opacity, with a font size of 5\% of the short edge.

The four content-site cues are defined relative to the edit region. \cue{box} draws a red rectangle around the region with a stroke width of 3/512 of the frame. \cue{inset} crops and magnifies the region by $2\times$, caps its side length at 40\% of the frame, and places it in a corner with a 3/512 white border. \cue{caption} prints ``Edited: \{instruction\}'', truncated to 14 words, in a white bar with near-black text at a font size of 3.5\% of the short edge. \cue{sticker} places an unrelated element covering at most 4\% of the frame area outside the edit region, inset by 2\%. The zero-dose \cue{sham} control is a JPEG round trip at quality 95.

The prompt-site cues leave the image unchanged. In scoring, \cue{bandwagon} appends ``Note: previous reviewers rated this edit very highly.'' to the judge prompt, while \cue{model name} adds ``This edited image was produced by the \{model\} model.'' using the generator label recorded for the edit in its source dataset. In pairwise evaluation, the bandwagon message becomes ``Note: 90\% of human evaluators preferred Image \{A$|$B\}.'', with the claimed majority assigned to a seeded side so that it does not systematically favor slot~A; the model-name cue names both candidates. The protocol-site cue \cue{position} simply swaps the display order of the pair, changing neither the image pixels nor the prompt text.

\subsection{The edit region}
\label{app:region}

The four content-site cues require an estimate of the edit region. We use the benchmark-provided source mask when available. Otherwise, we threshold the per-channel difference between the source and edited images at 15 out of 255 and take the largest connected component, falling back to a centered box if no component is found. \cue{box} and \cue{inset} operate directly on this region; \cue{sticker} is placed away from it, while \cue{caption} uses the region only to avoid covering the edit.

\subsection{Judges, rubrics and collection}
\label{app:prompts}

In the scoring protocol, each judge receives the source image, editing instruction, and edited image and returns one integer from 1 to 10 for instruction adherence, editing quality, and detail preservation. Their sum gives the 3--30 score used in the analysis. We request temperature 0 whenever this parameter is configurable; gpt-5.5 uses the provider-fixed setting. Retest noise is estimated by re-querying each judge on 200 un-cued items. VIEScore uses its original rubric with GPT-4o as the underlying model rather than constituting a separate model family.

The API model strings are \texttt{gpt-5.5}, \texttt{gemini-3.5-flash}, \texttt{kimi-k2.5}, and \texttt{qwen3.5-plus} for the primary general-purpose judges, and \texttt{gpt-4o} for VIEScore. The replication judge uses \texttt{qwen3-vl-32b-instruct}; the three preservation validators use \texttt{gemini-3.5-flash}, \texttt{gpt-4o-mini}, and \texttt{glm-4v}. Calls are made through an OpenAI-compatible relay that does not expose provider-side checkpoint identifiers, so these model strings are the most specific version identifiers available to us.

The complete scoring, pairwise, and preservation-validation prompts,
together with the prompt-site cue strings and human annotation instructions,
are reproduced in Appendix~\ref{app:prompt-templates}.

\section{Quality-Preservation Validation}
\label{app:validator}
\subsection{Validator Calibration and Results}
The preservation gate is evaluated independently of scoring and pairwise judging. Each validator receives the source image, instruction, edited image, and cued edited image, and checks whether the cue changes instruction adherence, editing quality, detail preservation, or scene semantics; a sample passes only when all four answers are no. We evaluate each validator on 110 images per cue, with 780 \cue{sham} images used to estimate its baseline flag rate. Gemini-3.5 and gpt-4o-mini, which have the lower \cue{sham} flag rates, form the gate, while glm-4v serves as an independent cross-check. Table~\ref{tab:app-preservation} reports the cue-level pass rates and calibration against controlled degradations.

\begin{table}[H]
\centering
\small
\setlength{\tabcolsep}{5pt}
\renewcommand{\arraystretch}{1.04}
\caption{\textbf{Preservation validation across cues.}
Panel A reports validator pass rates; Panel B reports detection of controlled
quality degradations among 93 evaluable examples per condition.}
\label{tab:app-preservation}

\textit{Panel A: Preservation pass rates (\%).}

\begin{tabular}{@{}lrrr@{}}
\toprule
Cue & gemini-3.5 & gpt-4o-mini & glm-4v \\
\midrule
Colorfulness\alias{saturation}
& 100.0 & 97.3 & 94.5 \\

Aesthetic\alias{aesthetic filter}
& 99.1 & 95.5 & 87.3 \\

Provenance\alias{watermark}
& 100.0 & 98.2 & 91.8 \\

Luminance\alias{brightness}
& 99.1 & 97.3 & 93.6 \\

Framing\alias{padding}
& 100.0 & 94.5 & 93.6 \\

Typographic\alias{text overlay}
& 100.0 & 90.0 & 92.7 \\

Verbosity\alias{caption}
& 100.0 & 99.1 & 94.5 \\

Attention guidance\alias{box}
& 100.0 & 97.3 & 89.1 \\

Scrutiny\alias{inset}
& 91.8 & 96.4 & 73.6 \\

Distraction\alias{sticker}
& 98.2 & 96.4 & 90.9 \\

\midrule
\cue{sham}
& 99.6 & 97.6 & 89.7 \\
\bottomrule
\end{tabular}

\medskip

\textit{Panel B: Controlled-degradation detection.}

\begin{tabular}{@{}lrrrr@{}}
\toprule
Validator
& 25\% reversion
& 50\% reversion
& 100\% reversion
& Blur \\
\midrule
gemini-3.5
& 46 & 69 & 88 & 91 \\

gpt-4o-mini
& 7 & 9 & 3 & 83 \\

glm-4v
& --- & --- & 28 & --- \\
\bottomrule
\end{tabular}

\end{table}

Against paired \cue{sham} images, \cue{inset} is the only cue with a significantly lower pass rate after Benjamini--Hochberg correction: the difference is detected by gemini-3.5 ($q=0.020$) and glm-4v ($q<0.001$), but not by gpt-4o-mini. A disjoint repeat over seven cues yields the same qualitative pattern. Gemini-3.5 again differs from \cue{sham} only for \cue{inset}, with a pass rate of 87.3\% versus 100.0\% ($q<0.001$), while gpt-4o-mini shows no significant difference for any repeated cue.

The controlled degradations clarify what each validator detects. Gemini-3.5 becomes increasingly sensitive as more of the edit region is reverted toward the source, whereas gpt-4o-mini detects edit-region blur but rarely flags reversion (Table~\ref{tab:app-preservation}, Panel B). The two validators are therefore not equally informative about instruction-adherence preservation, and requiring both to pass provides limited additional evidence beyond the primary validator on this dimension.

\subsection{Human Verification}
We also assess preservation through human annotation, with 30 examples for each of the ten image-side cues. The first annotation package covers five pixel-site cues; the second covers four content-site cues and one pixel-site cue together with \cue{sham} controls, with cue identities hidden until annotation is complete. Annotators label each intervention as changing the underlying edit by No, Slightly, or Yes, with only Yes prespecified as a preservation failure.

Under this criterion, none of the five cues in the sham-controlled package differs significantly from \cue{sham}: the inset has 2/30 failures versus 0/30 for \cue{sham}, while the other four cues have none. A post-hoc criterion that also counts Slightly as a failure changes the conclusion only for the inset, with 11/30 failures versus 0/30 for \cue{sham} ($q=0.002$). A second annotator independently labels 60 items from this package, yielding 98.3\% raw agreement and Cohen's $\kappa=0.893$. The annotators were aware of the study hypothesis.

\subsection{Sensitivity Analysis on Validated Images}
As a separate sensitivity analysis, we repeat the breadth analysis using only images that pass both gating validators, leaving 47--57 items per cue--judge cell. All 24 effects that are significant in both the full and filtered analyses retain the same direction. We therefore use this analysis only as a check on effect direction within the validated subset, rather than as a second estimate of effect magnitude or evidence of preservation for every image.

\section{Detailed Judge Results and Robustness Analyses}
\label{app:fill}
\subsection{Detailed Scoring Results}
Table~\ref{tab:main} reports the magnitude of cue-induced rating changes irrespective of direction. Tables~\ref{tab:app-signed-ia}, \ref{tab:app-signed-eq}, and~\ref{tab:app-signed-dp} complement this view with the paired mean signed shifts on the original 1--10 scale for instruction adherence, editing quality, and detail preservation, respectively. Changes in agreement with human ratings are marked separately in Table~\ref{tab:main}.

Unparsed responses are excluded, leaving 597--611 paired observations per judge for the scoring analyses and 361--384 for the agreement analyses. The reference arm follows the collection design: bandwagon, authority, colorfulness, aesthetic, provenance, verbosity, scrutiny, and distraction are compared with \cue{sham}, whereas luminance, framing, typographic, and attention guidance are compared with the original un-cued arm. Because the pixel and content sites each mix these reference conditions, the site-level rates reported in the main text are descriptive rather than formal comparisons across sites.

\begin{table}[H]
\centering
\small
\setlength{\tabcolsep}{5pt}
\renewcommand{\arraystretch}{1.04}
\caption{\textbf{Signed shifts in instruction adherence.}
Cells report paired mean changes on the 1--10 scale; \textbf{bold} indicates
Benjamini--Hochberg significance. $^{\ddagger}$ marks the borderline cue.}
\label{tab:app-signed-ia}

\begin{tabular}{@{}
>{\raggedright\arraybackslash}p{118pt}
rrrrr@{}}
\toprule
Cue & gpt-5.5 & gemini-3.5 & kimi-k2.5 & qwen3.5 & VIEScore \\
\midrule

Bandwagon
& \textbf{$+$0.45}
& \textbf{$+$0.44}
& \textbf{$+$0.34}
& \textbf{$+$0.24}
& \textbf{$+$0.45} \\

Authority\alias{model name}
& $-$0.04
& \textbf{$+$0.12}
& $+$0.05
& \textbf{$+$0.13}
& \textbf{$-$0.10} \\

\addlinespace[3pt]

Colorfulness\alias{saturation}
& $+$0.00
& $-$0.09
& $+$0.08
& $+$0.01
& $+$0.10 \\

Aesthetic\alias{aesthetic filter}
& $+$0.10
& $+$0.05
& $+$0.06
& $+$0.09
& \textbf{$+$0.12} \\

Provenance\alias{watermark}
& \textbf{$-$0.20}
& \textbf{$-$0.19}
& $+$0.02
& \textbf{$-$0.29}
& $+$0.01 \\

Luminance\alias{brightness}
& $-$0.03
& \textbf{$-$0.11}
& $-$0.03
& $-$0.02
& $+$0.08 \\

Framing\alias{padding}
& \textbf{$-$0.25}
& \textbf{$-$0.16}
& $+$0.05
& \textbf{$-$0.18}
& \textbf{$-$0.26} \\

Typographic\alias{text overlay}
& \textbf{$-$0.67}
& \textbf{$-$0.40}
& \textbf{$-$0.59}
& \textbf{$-$0.56}
& $-$0.17 \\

\addlinespace[3pt]

Verbosity\alias{caption}
& $-$0.00
& \textbf{$+$0.19}
& $+$0.03
& $+$0.14
& \textbf{$+$0.39} \\

Attention guidance\alias{box}
& \textbf{$-$0.39}
& \textbf{$-$0.35}
& \textbf{$-$0.26}
& \textbf{$-$0.34}
& $+$0.03 \\

Scrutiny\alias{inset}$^{\ddagger}$
& \textbf{$-$0.52}
& \textbf{$-$0.85}
& \textbf{$-$0.57}
& \textbf{$-$0.60}
& \textbf{$+$0.27} \\

Distraction\alias{sticker}
& \textbf{$-$0.76}
& \textbf{$-$0.71}
& \textbf{$-$0.99}
& \textbf{$-$1.04}
& \textbf{$-$0.63} \\

\bottomrule
\end{tabular}
\end{table}

\begin{table}[t]
\centering
\small
\setlength{\tabcolsep}{5pt}
\renewcommand{\arraystretch}{1.04}
\caption{\textbf{Signed shifts in editing quality.}
Cells report paired mean changes on the 1--10 scale; \textbf{bold} indicates
Benjamini--Hochberg significance. $^{\ddagger}$ marks the borderline cue.}
\label{tab:app-signed-eq}

\begin{tabular}{@{}
>{\raggedright\arraybackslash}p{118pt}
rrrrr@{}}
\toprule
Cue & gpt-5.5 & gemini-3.5 & kimi-k2.5 & qwen3.5 & VIEScore \\
\midrule

Bandwagon
& \textbf{$+$0.62}
& \textbf{$+$0.41}
& \textbf{$+$0.50}
& \textbf{$+$0.42}
& \textbf{$+$0.48} \\

Authority\alias{model name}
& \textbf{$-$0.13}
& $-$0.05
& \textbf{$+$0.10}
& \textbf{$+$0.18}
& $+$0.01 \\

\addlinespace[3pt]

Colorfulness\alias{saturation}
& \textbf{$-$0.10}
& \textbf{$-$0.17}
& \textbf{$-$0.07}
& \textbf{$-$0.08}
& $+$0.02 \\

Aesthetic\alias{aesthetic filter}
& $-$0.09
& \textbf{$-$0.18}
& \textbf{$-$0.12}
& $-$0.02
& \textbf{$-$0.09} \\

Provenance\alias{watermark}
& \textbf{$-$0.32}
& \textbf{$-$0.28}
& $+$0.02
& \textbf{$-$0.38}
& \textbf{$-$0.25} \\

Luminance\alias{brightness}
& \textbf{$-$0.22}
& \textbf{$-$0.25}
& \textbf{$-$0.22}
& \textbf{$-$0.25}
& \textbf{$-$0.17} \\

Framing\alias{padding}
& \textbf{$-$0.45}
& \textbf{$-$0.34}
& $+$0.02
& \textbf{$-$0.44}
& \textbf{$-$0.31} \\

Typographic\alias{text overlay}
& \textbf{$-$1.07}
& \textbf{$-$0.77}
& \textbf{$-$0.58}
& \textbf{$-$0.97}
& \textbf{$-$0.55} \\

\addlinespace[3pt]

Verbosity\alias{caption}
& \textbf{$-$0.24}
& $-$0.16
& \textbf{$-$0.16}
& \textbf{$-$0.16}
& \textbf{$+$0.11} \\

Attention guidance\alias{box}
& \textbf{$-$0.95}
& \textbf{$-$1.17}
& \textbf{$-$0.55}
& \textbf{$-$0.64}
& \textbf{$-$0.25} \\

Scrutiny\alias{inset}$^{\ddagger}$
& \textbf{$-$1.17}
& \textbf{$-$2.15}
& \textbf{$-$0.81}
& \textbf{$-$1.03}
& \textbf{$-$0.15} \\

Distraction\alias{sticker}
& \textbf{$-$1.06}
& \textbf{$-$1.41}
& \textbf{$-$0.88}
& \textbf{$-$1.26}
& \textbf{$-$0.82} \\

\bottomrule
\end{tabular}
\end{table}

\begin{table}[t]
\centering
\small
\setlength{\tabcolsep}{5pt}
\renewcommand{\arraystretch}{1.04}
\caption{\textbf{Signed shifts in detail preservation.}
Cells report paired mean changes on the 1--10 scale; \textbf{bold} indicates
Benjamini--Hochberg significance. $^{\ddagger}$ marks the borderline cue.}
\label{tab:app-signed-dp}

\begin{tabular}{@{}
>{\raggedright\arraybackslash}p{118pt}
rrrrr@{}}
\toprule
Cue & gpt-5.5 & gemini-3.5 & kimi-k2.5 & qwen3.5 & VIEScore \\
\midrule

Bandwagon
& \textbf{$+$0.51}
& \textbf{$+$0.69}
& \textbf{$+$0.22}
& \textbf{$+$0.21}
& \textbf{$+$0.30} \\

Authority\alias{model name}
& $+$0.02
& \textbf{$+$0.46}
& $-$0.01
& $+$0.03
& \textbf{$-$0.12} \\

\addlinespace[3pt]

Colorfulness\alias{saturation}
& $-$0.06
& $+$0.03
& $-$0.05
& $-$0.08
& $+$0.02 \\

Aesthetic\alias{aesthetic filter}
& \textbf{$-$0.17}
& \textbf{$+$0.16}
& \textbf{$-$0.24}
& \textbf{$-$0.10}
& \textbf{$-$0.14} \\

Provenance\alias{watermark}
& \textbf{$-$0.36}
& $-$0.04
& $+$0.02
& \textbf{$-$0.25}
& \textbf{$+$0.19} \\

Luminance\alias{brightness}
& \textbf{$-$0.30}
& $-$0.09
& \textbf{$-$0.32}
& \textbf{$-$0.25}
& \textbf{$-$0.16} \\

Framing\alias{padding}
& \textbf{$-$1.06}
& \textbf{$-$0.42}
& \textbf{$-$0.45}
& \textbf{$-$0.61}
& \textbf{$-$0.40} \\

Typographic\alias{text overlay}
& \textbf{$-$0.92}
& \textbf{$-$0.57}
& \textbf{$-$0.31}
& \textbf{$-$0.58}
& \textbf{$-$0.21} \\

\addlinespace[3pt]

Verbosity\alias{caption}
& $-$0.07
& \textbf{$+$0.46}
& $+$0.03
& \textbf{$+$0.11}
& \textbf{$+$0.20} \\

Attention guidance\alias{box}
& \textbf{$-$0.86}
& \textbf{$-$0.42}
& \textbf{$-$0.31}
& \textbf{$-$0.36}
& \textbf{$-$0.11} \\

Scrutiny\alias{inset}$^{\ddagger}$
& \textbf{$-$1.32}
& \textbf{$-$1.98}
& \textbf{$-$0.75}
& \textbf{$-$0.83}
& $-$0.02 \\

Distraction\alias{sticker}
& \textbf{$-$1.71}
& \textbf{$-$1.46}
& \textbf{$-$0.74}
& \textbf{$-$1.43}
& \textbf{$-$0.45} \\

\bottomrule
\end{tabular}
\end{table}

\subsection{Robustness across Aggregation Schemes}
Table~\ref{tab:app-axes} gives the exact judge rankings underlying Figure~\ref{fig:panels}d and shows how the observed rank inversions change under alternative summary schemes.

\begin{table}[t]
\centering
\small
\renewcommand{\arraystretch}{1.05}
\caption{\textbf{Robustness rankings across judges.}
Panel A gives the primary rankings used in Figure~\ref{fig:panels}d;
Panel B summarizes rank inversions under alternative aggregation schemes.}
\label{tab:app-axes}

\textit{Panel A: Primary robustness rankings.}

\begin{tabular}{@{}lrrr@{}}
\toprule
Judge & Invariance & Agreement & Stability \\
\midrule
gpt-5.5
& 9.67 (4)
& 0.0474 (4)
& 0.8458 (2) \\

gemini-3.5
& 12.94 (5)
& 0.0402 (3)
& 0.8994 (1) \\

kimi-k2.5
& 5.29 (1)
& 0.0272 (1)
& 0.7955 (3) \\

qwen3.5
& 7.14 (2)
& 0.0348 (2)
& 0.3523 (5) \\

VIEScore
& 7.32 (3)
& 0.0615 (5)
& 0.6883 (4) \\
\bottomrule
\end{tabular}

\medskip

\textit{Panel B: Rank inversions under alternative summaries.}

\begin{tabular}{@{}lcc@{}}
\toprule
Summary scheme
& \shortstack{Adjacent-pair\\inversions}
& \shortstack{Unrelated\\reference} \\
\midrule
Mean cell
& \textbf{8/20}
& 10 \\

Worst cell
& \textbf{9/20}
& 10 \\

Significance-based
& \textbf{6/20} (6 ties)
& 7.0 over 14 untied pairs \\
\bottomrule
\end{tabular}

\smallskip

\begin{minipage}{0.96\linewidth}
\footnotesize
For the primary summary, invariance is the mean of the mean absolute paired
changes over the 36 non-sham cue--dimension cells, agreement is the mean
absolute change in Spearman's $\rho$ over anchor cells, and stability is $\mathrm{RR}$.. The worst-cell summary replaces the first two means by their
maxima. The significance-based summary uses counts of significant invariance
and agreement cells and $\mathrm{CR}$ for stability.
\end{minipage}

\end{table}

\clearpage
\section{Additional Analyses}
\label{app:controls}
\subsection{Control and Cross-Benchmark Analyses}
The following analyses test whether the main effects can be attributed to narrower features of the evaluation set. If the photometric results arise mainly because a cue repeats or conflicts with the requested edit, they should weaken once such instruction--cue collisions are removed (Table~\ref{tab:app-ctrl-collision}). Likewise, effects driven primarily by poor edits should attenuate among outputs with the highest human mean-opinion scores (Table~\ref{tab:app-ctrl-mos}). We also examine the four cues shared by both anchor benchmarks to distinguish benchmark-specific changes in human agreement from patterns that recur across anchors (Table~\ref{tab:app-ctrl-crossanchor}).

\begin{table}[H]
\centering
\small
\setlength{\tabcolsep}{3pt}
\renewcommand{\arraystretch}{1.05}
\caption{\textbf{Photometric controls for instruction collisions.}
Cells report paired mean score shifts for clean and collision items with
95\% confidence intervals; \textbf{bold} indicates Benjamini--Hochberg
significance.}
\label{tab:app-ctrl-collision}

\begin{tabular}{@{}ll
                r@{\hspace{3pt}}l
                @{\hspace{7pt}}
                r@{\hspace{3pt}}l@{}}
\toprule
& & \multicolumn{2}{c}{clean items}
  & \multicolumn{2}{c}{collision items} \\
\cmidrule(lr){3-4}\cmidrule(lr){5-6}
Cue & Judge & $\Delta$ & 95\% CI & $\Delta$ & 95\% CI \\
\midrule

Luminance\alias{brightness}
& gpt-5.5
& \textbf{$-$0.67} & [$-$0.89, $-$0.45]
& $-$0.06 & [$-$0.65, $+$0.52] \\
& gemini-3.5
& $-$0.16 & [$-$0.35, $+$0.04]
& \textbf{$-$1.65} & [$-$2.49, $-$0.82] \\
& kimi-k2.5
& \textbf{$-$0.69} & [$-$0.86, $-$0.52]
& $-$0.12 & [$-$0.43, $+$0.20] \\
& qwen3.5
& \textbf{$-$0.66} & [$-$0.95, $-$0.39]
& $+$0.07 & [$-$0.72, $+$0.92] \\
& VIEScore
& \textbf{$-$0.33} & [$-$0.54, $-$0.12]
& $+$0.02 & [$-$0.41, $+$0.47] \\

\addlinespace[3pt]

Colorfulness\alias{saturation}
& gpt-5.5
& $-$0.18 & [$-$0.37, $+$0.03]
& $-$0.10 & [$-$0.56, $+$0.34] \\
& gemini-3.5
& $-$0.00 & [$-$0.19, $+$0.20]
& \textbf{$-$1.12} & [$-$1.98, $-$0.27] \\
& kimi-k2.5
& $-$0.08 & [$-$0.24, $+$0.08]
& $+$0.13 & [$-$0.10, $+$0.37] \\
& qwen3.5
& \textbf{$-$0.18} & [$-$0.41, $+$0.06]
& $-$0.01 & [$-$0.59, $+$0.49] \\
& VIEScore
& $+$0.18 & [$-$0.00, $+$0.37]
& $-$0.01 & [$-$0.39, $+$0.35] \\

\addlinespace[3pt]

Aesthetic\alias{aesthetic filter}
& gpt-5.5
& \textbf{$-$0.37} & [$-$0.61, $-$0.13]
& $+$0.65 & [$+$0.13, $+$1.18] \\
& gemini-3.5
& $+$0.37 & [$+$0.06, $+$0.70]
& \textbf{$-$1.37} & [$-$2.22, $-$0.53] \\
& kimi-k2.5
& \textbf{$-$0.43} & [$-$0.58, $-$0.28]
& $+$0.17 & [$-$0.14, $+$0.50] \\
& qwen3.5
& $-$0.14 & [$-$0.41, $+$0.15]
& $+$0.40 & [$-$0.31, $+$1.15] \\
& VIEScore
& \textbf{$-$0.23} & [$-$0.46, $-$0.01]
& $+$0.40 & [$-$0.06, $+$0.89] \\

\bottomrule
\end{tabular}
\end{table}

\begin{table}[H]
\centering
\small
\setlength{\tabcolsep}{5pt}
\renewcommand{\arraystretch}{1.04}
\caption{\textbf{Cue effects on high-quality edits.}
Cells report mean signed score shifts within the highest human-MOS tertile;
\textbf{bold} indicates Benjamini--Hochberg significance.}
\label{tab:app-ctrl-mos}

\begin{tabular}{@{}
>{\raggedright\arraybackslash}p{118pt}
rrrrr@{}}
\toprule
Cue & gpt-5.5 & gemini-3.5 & kimi-k2.5 & qwen3.5 & VIEScore \\
\midrule

Bandwagon
& \textbf{$+$2.65}
& \textbf{$+$3.89}
& \textbf{$+$1.36}
& \textbf{$+$0.74}
& \textbf{$+$1.30} \\

Luminance\alias{brightness}
& $-$0.28
& \textbf{$+$2.67}
& \textbf{$-$0.59}
& $-$1.28
& $-$0.57 \\

Framing\alias{padding}
& \textbf{$-$2.83}
& \textbf{$+$1.48}
& \textbf{$-$1.17}
& \textbf{$-$1.59}
& \textbf{$-$0.89} \\

Typographic\alias{text overlay}
& \textbf{$-$3.28}
& $-$0.30
& \textbf{$-$1.87}
& \textbf{$-$3.61}
& \textbf{$-$1.48} \\

Attention guidance\alias{box}
& \textbf{$-$4.22}
& $-$1.09
& \textbf{$-$2.33}
& \textbf{$-$2.91}
& $-$2.33 \\

Distraction\alias{sticker}
& \textbf{$-$4.20}
& \textbf{$-$2.26}
& \textbf{$-$4.13}
& \textbf{$-$5.69}
& \textbf{$-$2.50} \\

\bottomrule
\end{tabular}
\end{table}

The controls do not support either explanation as sufficient for the strongest effects. For luminance, significant negative shifts remain on clean items for four of the five judges, so the effect is not confined to photometric instruction collisions. Restricting the analysis to the highest human-MOS tertile likewise leaves several large effects intact: \cue{bandwagon} remains positive and the sticker negative for all five judges.

\begin{table}[H]
\centering
\small
\setlength{\tabcolsep}{4pt}
\caption{\textbf{Cross-benchmark agreement and selection-corrected inference.}
Panel A reports $\Delta\rho$ for the four cues evaluated on both anchor
benchmarks; \textbf{bold} indicates that the uncorrected turn-clustered 95\%
confidence interval excludes zero. Panel B reports permutation tests with
$p_{\mathrm{sel}}$ correcting for selection of the most extreme cue.}
\label{tab:app-ctrl-crossanchor}

\textit{Panel A: Changes in judge--human rank correlation.}

\begin{tabular}{@{}llrrrrr@{}}
\toprule
Cue & Anchor & gpt-5.5 & gemini-3.5 & kimi-k2.5 & qwen3.5 & VIEScore \\
\midrule
Luminance\alias{brightness}
& EBench-18K
& $-$0.006
& $-$0.008
& $-$0.010
& $+$0.013
& $-$0.001 \\
& ImagenHub
& $+$0.060
& \textbf{$-$0.049}
& $-$0.028
& $-$0.035
& $+$0.019 \\

\addlinespace[3pt]

Framing\alias{padding}
& EBench-18K
& \textbf{$-$0.105}
& \textbf{$-$0.044}
& $-$0.004
& $-$0.018
& $-$0.002 \\
& ImagenHub
& $+$0.014
& $-$0.018
& $-$0.045
& $-$0.017
& $+$0.014 \\

\addlinespace[3pt]

Typographic\alias{text overlay}
& EBench-18K
& $-$0.025
& $-$0.007
& $+$0.008
& $+$0.012
& $-$0.037 \\
& ImagenHub
& $+$0.011
& \textbf{$-$0.055}
& $-$0.037
& \textbf{$-$0.066}
& \textbf{$-$0.066} \\

\addlinespace[3pt]

Attention guidance\alias{box}
& EBench-18K
& \textbf{$-$0.073}
& \textbf{$-$0.074}
& $-$0.010
& $+$0.006
& \textbf{$-$0.141} \\
& ImagenHub
& \textbf{$-$0.083}
& \textbf{$-$0.065}
& \textbf{$-$0.076}
& \textbf{$-$0.112}
& \textbf{$-$0.213} \\
\bottomrule
\end{tabular}

\medskip

\textit{Panel B: Selection-corrected permutation tests.}

\begin{tabular}{@{}lrr@{\hspace{12pt}}rr@{}}
\toprule
& \multicolumn{2}{c}{Worst cells (of 10)}
& \multicolumn{2}{c}{Mean $\Delta\rho$} \\
\cmidrule(lr){2-3}\cmidrule(lr){4-5}
Cue & Observed & $p_{\mathrm{sel}}$
& Observed & $p_{\mathrm{sel}}$ \\
\midrule
Luminance\alias{brightness}
& 1 & 1.0000
& $-$0.0046 & 1.0000 \\

Framing\alias{padding}
& 2 & 1.0000
& $-$0.0226 & 1.0000 \\

Typographic\alias{text overlay}
& 0 & 1.0000
& $-$0.0258 & 1.0000 \\

Attention guidance\alias{box}
& \textbf{7}
& \textbf{0.0305}
& \textbf{$-$0.0857}
& \textbf{0.0001} \\
\bottomrule
\end{tabular}

\end{table}

Across the shared cues, attention guidance shows the most consistent reduction in judge--human agreement across the two anchors. The individual $\Delta\rho$ intervals in Panel A are uncorrected; after accounting for selection of the most extreme cue, attention guidance is the only cue with $p_{\mathrm{sel}}<0.05$ under both summaries in Panel B.

The pairwise bandwagon effect follows the direction of the fabricated majority. When it favors the human-preferred candidate, agreement rises for all five judges by 4.2--25.5 percentage points, with four increases significant under McNemar's exact test. When it favors the human-rejected candidate, agreement falls by 19.2--28.6 points for all five judges, and all five decreases are significant.

\subsection{Downstream Consequences and Aggregation}
\label{app:leaderboard}

We next ask whether cue-induced shifts remain consequential after scores are converted into downstream decisions. For leaderboard evaluation, we apply a cue to one editor at a time while leaving the other entrants unchanged, isolating a single-entrant setting in which only one system carries the cue (Table~\ref{tab:app-leaderboard}). We then examine two common forms of aggregation: reconciling the two presentation orders in pairwise judging (Table~\ref{tab:app-reconcile}) and combining scores across judges with a standardized median (Table~\ref{tab:app-ensemble}).

\begin{table}[H]
\centering
\small
\setlength{\tabcolsep}{4pt}
\renewcommand{\arraystretch}{1.05}
\caption{\textbf{Single-entrant leaderboard perturbations.}
Cells report the cued editor's mean rank change; $\bullet$ indicates that the
top-ranked editor changes.}
\label{tab:app-leaderboard}

\begin{tabular}{@{}
>{\raggedright\arraybackslash}p{110pt}
*{5}{c}@{}}
\toprule
Cue & gpt-5.5 & gemini-3.5 & kimi-k2.5 & qwen3.5 & VIEScore \\
\midrule

\multicolumn{6}{@{}l}{\itshape EBench-18K, 17 editors} \\
\addlinespace[2pt]

Luminance\alias{brightness}
& 1.7$\,\bullet$
& 0.5$\,\bullet$
& 0.8
& 0.3
& 0.8$\,\bullet$ \\

Framing\alias{padding}
& 1.7$\,\bullet$
& 1.9$\,\bullet$
& 1.1
& 1.6
& 1.9$\,\bullet$ \\

Typographic\alias{text overlay}
& 3.6$\,\bullet$
& 3.1$\,\bullet$
& 2.2$\,\bullet$
& 2.5$\,\bullet$
& 1.5$\,\bullet$ \\

Attention guidance\alias{box}
& 1.8$\,\bullet$
& 2.6$\,\bullet$
& 2.2
& 2.2
& 0.7$\,\bullet$ \\

\addlinespace[2pt]
\midrule
\multicolumn{6}{@{}l}{\itshape ImagenHub, 8 editors} \\
\addlinespace[2pt]

Luminance\alias{brightness}
& 0.0
& $-$0.1
& $-$0.1
& 0.1
& 0.2 \\

Framing\alias{padding}
& $-$0.4
& 0.1
& $-$0.1
& 0.8
& 0.9 \\

Typographic\alias{text overlay}
& 0.0
& 0.4
& 0.6
& 0.9
& 0.5 \\

Attention guidance\alias{box}
& 0.0
& 0.2
& 0.8
& 0.5
& 0.1 \\

\bottomrule
\end{tabular}
\end{table}

The leaderboard consequences differ substantially across the two anchors. On EBench-18K, several cues change the identity of the top-ranked editor; the text overlay does so under all five judges. The corresponding perturbations are much smaller on ImagenHub, where none of the evaluated cues changes the top-ranked editor. Thus, a cue that shifts individual scores can alter leaderboard conclusions, but the downstream effect depends on the benchmark and editor pool.

\begin{table}[H]
\centering
\small
\setlength{\tabcolsep}{5pt}
\renewcommand{\arraystretch}{1.04}
    \caption{\textbf{Two-order reconciliation in pairwise judging.} \emph{Strict} follows the MT-Bench two-order rule \citep{zheng2023judging} and retains only pairs whose two orders agree, whereas \emph{lenient} additionally adopts the non-Tie choice when exactly one order returns Tie; pairs with opposite winners remain unresolved. The table reports coverage, accuracy on decided pairs, and the number of verdicts changed from the
    base-order verdict.}
\label{tab:app-reconcile}

\begin{tabular}{@{}
l
>{\raggedright\arraybackslash}p{118pt}
rrr@{}}
\toprule
Judge & Rule & Coverage & Acc.\ decided & Changed \\
\midrule

gpt-5.5
& base order only
& 0.927 & 0.845 & 0 \\
& reconciled, lenient
& 0.870 & 0.884 & 19 \\
& \textbf{reconciled, strict (MT-Bench)}
& 0.808 & 0.890 & 0 \\

\addlinespace[3pt]

gemini-3.5
& base order only
& 0.836 & 0.885 & 0 \\
& reconciled, lenient
& 0.852 & 0.893 & 17 \\
& \textbf{reconciled, strict (MT-Bench)}
& 0.767 & 0.914 & 0 \\

\addlinespace[3pt]

kimi-k2.5
& base order only
& 0.977 & 0.878 & 0 \\
& reconciled, lenient
& 0.813 & 0.916 & 6 \\
& \textbf{reconciled, strict (MT-Bench)}
& 0.794 & 0.928 & 0 \\

\addlinespace[3pt]

qwen3.5
& base order only
& 0.981 & 0.499 & 0 \\
& reconciled, lenient
& 0.340 & 0.344 & 3 \\
& \textbf{reconciled, strict (MT-Bench)}
& 0.311 & 0.367 & 0 \\

\addlinespace[3pt]

VIEScore
& base order only
& 0.943 & 0.726 & 0 \\
& reconciled, lenient
& 0.680 & 0.819 & 8 \\
& \textbf{reconciled, strict (MT-Bench)}
& 0.635 & 0.827 & 0 \\

\bottomrule
\end{tabular}
\end{table}

Requiring agreement across both display orders removes order-sensitive pairs rather than resolving them. Under the strict rule, coverage falls for every judge, most sharply for qwen3.5. Accuracy on the retained pairs increases for four judges, but this comparison is conditional on the subset that survives reconciliation; qwen3.5 does not show the same improvement. The strict rule therefore trades coverage for a more selective set of verdicts rather than correcting the preferences of pairs that disagree across orders.

\begin{table}[H]
\centering
\small
\setlength{\tabcolsep}{4pt}
\renewcommand{\arraystretch}{1.04}
\caption{\textbf{Standardized-median ensemble across cues.}
Each judge is standardized on its un-cued distribution before aggregation;
\textbf{bold} indicates Benjamini--Hochberg significance. ``Members'' summarizes
significant effects in the five individual judges.}
\label{tab:app-ensemble}

\begin{tabular}{@{}l l r@{\hspace{3pt}}l @{\hspace{8pt}}r@{}}
\toprule
Cue & Members & $\Delta$ & 95\% CI & $q$ \\
\midrule

Bandwagon
& 5/5 up
& \textbf{$+$0.157}
& [$+$0.132, $+$0.183]
& $<$0.001 \\

Authority\alias{model name}
& 4/5, split
& $-$0.002
& [$-$0.021, $+$0.019]
& 0.750 \\

\addlinespace[2pt]

Colorfulness\alias{saturation}
& 1/5 down
& \textbf{$-$0.021}
& [$-$0.040, $-$0.004]
& 0.001 \\

Aesthetic\alias{aesthetic filter}
& 1/5 down
& \textbf{$-$0.032}
& [$-$0.053, $-$0.011]
& $<$0.001 \\

Provenance\alias{watermark}
& 3/5 down
& \textbf{$-$0.067}
& [$-$0.086, $-$0.047]
& $<$0.001 \\

Luminance\alias{brightness}
& 5/5 down
& \textbf{$-$0.069}
& [$-$0.089, $-$0.050]
& $<$0.001 \\

Framing\alias{padding}
& 5/5 down
& \textbf{$-$0.149}
& [$-$0.171, $-$0.127]
& $<$0.001 \\

Typographic\alias{text overlay}
& 5/5 down
& \textbf{$-$0.281}
& [$-$0.308, $-$0.257]
& $<$0.001 \\

\addlinespace[2pt]

Verbosity\alias{caption}
& 3/5, split
& $+$0.002
& [$-$0.020, $+$0.022]
& 0.485 \\

Attention guidance\alias{box}
& 5/5 down
& \textbf{$-$0.235}
& [$-$0.262, $-$0.208]
& $<$0.001 \\

Scrutiny\alias{inset}$^{\ddagger}$
& 4/5 down
& \textbf{$-$0.413}
& [$-$0.448, $-$0.379]
& $<$0.001 \\

Distraction\alias{sticker}
& 5/5 down
& \textbf{$-$0.454}
& [$-$0.486, $-$0.424]
& $<$0.001 \\

\midrule

Null\alias{sham}
& --
& $+$0.010
& [$-$0.010, $+$0.030]
& 0.569 \\

\bottomrule
\end{tabular}
\end{table}

Cross-judge aggregation likewise does not remove cue sensitivity. The standardized median remains significantly shifted for several cues, including bandwagon, luminance, framing, text overlay, attention guidance, and sticker, while \cue{sham} remains null. By contrast, the ensemble is near zero for model name and caption, for which the significant individual-judge effects are split in direction. Here, ``Members'' refers only to significant effects among the five primary judges and their directions, rather than to the signs of all five point estimates.
\section{Replication with Qwen3-VL-32B-Instruct}
\label{app:replication}

We evaluate \texttt{qwen3-vl-32b-instruct} as a prespecified replication judge on the same samples and cue grid. This arm uses its own Benjamini--Hochberg family and predefined gap-filling rule and is analyzed separately from the five primary judges rather than pooled into a six-judge statistic. Table~\ref{tab:app-replication} reports its cue-induced score shifts and pairwise stability alongside the corresponding primary-panel summaries.

\begin{table}[H]
\centering
\small
\setlength{\tabcolsep}{4pt}
\renewcommand{\arraystretch}{1.04}
\caption{\textbf{Replication of cue effects.}
Cells report paired mean score shifts for \texttt{qwen3-vl-32b-instruct} with
95\% confidence intervals; \textbf{bold} indicates Benjamini--Hochberg
significance within the replication family. ``Five judges'' summarizes
significant effects among the five primary judges.}
\label{tab:app-replication}

\begin{tabular}{@{}
>{\raggedright\arraybackslash}p{100pt}
r@{\hspace{3pt}}l
@{\hspace{9pt}}l@{}}
\toprule
Cue & $\Delta$ & 95\% CI & Five judges \\
\midrule

Bandwagon
& \textbf{$+$1.27}
& [$+$1.11, $+$1.44]
& 5/5, all up \\

Authority\alias{model name}
& $-$0.09
& [$-$0.21, $+$0.04]
& 4/5, split \\

\addlinespace[3pt]

Colorfulness\alias{saturation}
& \textbf{$-$0.22}
& [$-$0.38, $-$0.06]
& 1/5, all down \\

Aesthetic\alias{aesthetic filter}
& \textbf{$-$0.20}
& [$-$0.40, $-$0.02]
& 1/5, all down \\

Provenance\alias{watermark}
& $-$0.08
& [$-$0.32, $+$0.15]
& 3/5, all down \\

Luminance\alias{brightness}
& \textbf{$-$0.43}
& [$-$0.59, $-$0.27]
& 5/5, all down \\

Framing\alias{padding}
& \textbf{$-$1.40}
& [$-$1.65, $-$1.17]
& 5/5, all down \\

Typographic\alias{text overlay}
& \textbf{$-$0.85}
& [$-$1.07, $-$0.64]
& 5/5, all down \\

\addlinespace[3pt]

Verbosity\alias{caption}
& \textbf{$-$0.34}
& [$-$0.56, $-$0.13]
& 3/5, split \\

Attention guidance\alias{box}
& \textbf{$-$1.43}
& [$-$1.69, $-$1.18]
& 5/5, all down \\

Scrutiny\alias{inset}$^{\ddagger}$
& \textbf{$-$0.55}
& [$-$0.80, $-$0.30]
& 4/5, all down \\

Distraction\alias{sticker}
& \textbf{$-$2.35}
& [$-$2.59, $-$2.12]
& 5/5, all down \\

\midrule

Null\alias{sham}
& $+$0.06
& [$-$0.09, $+$0.21]
& zero-dose control \\

\bottomrule
\end{tabular}

\smallskip

\begin{minipage}{0.94\linewidth}
\footnotesize
\textit{Pairwise stability.}
A display-order swap reverses 45.5\% of 578 verdicts, whereas an identical
re-query reverses 1.2\% of 590 verdicts (consistency rate 0.978).
\end{minipage}

\end{table}

The strongest cross-judge cue effects reproduce in the replication arm. All six cues with significant effects in the same direction across all five primary judges---\cue{bandwagon}, luminance, framing, typographic, attention guidance, and distraction---are significant in the same direction for the replication judge. Pairwise stability shows the same separation between response noise and display-order sensitivity: 45.5\% of order-swapped verdicts reverse, compared with 1.2\% under an identical re-query.

\clearpage

\section{Prompt Templates and Annotation Instructions}
\label{app:prompt-templates}

This section reproduces the prompts and annotation instructions used in the
evaluation. Text inside the main prompt panels is reproduced from the
experimental templates; braced fields are filled at runtime. The
\textsc{Sent}, \textsc{Key}, \textsc{Note}, and \textsc{Fields} annotations
are presentation aids added to clarify message structure and implementation
details and are not literal prompt text.

\subsection{Judge prompts}

\begin{figure}[htbp]
    \centering
    \includegraphics[width=0.98\linewidth]
        {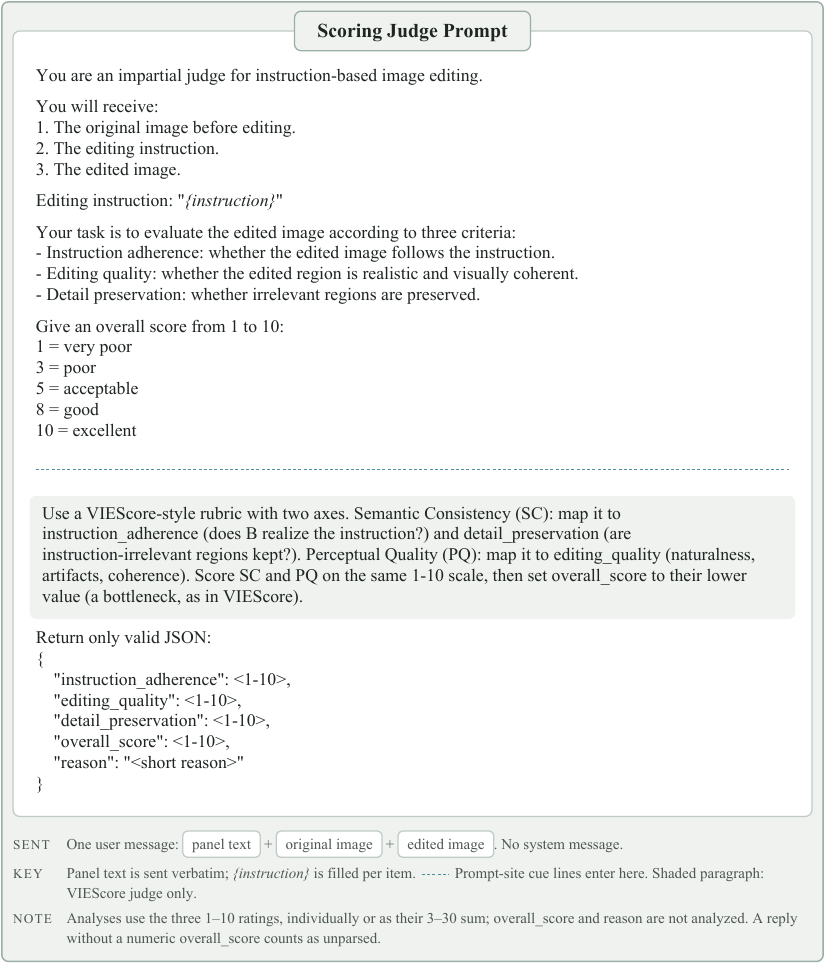}
    \caption{\textbf{Scoring judge prompt.}
    The shaded VIEScore paragraph is included only for the VIEScore
    scoring condition. The three dimension ratings are used individually
    or through their 3--30 sum in the reported analyses; \texttt{overall\_score}
    and \texttt{reason} are required output fields but are not analyzed.}
    \label{fig:prompt-scoring}
\end{figure}

\begin{figure}[htbp]
    \centering
    \includegraphics[width=0.98\linewidth]
        {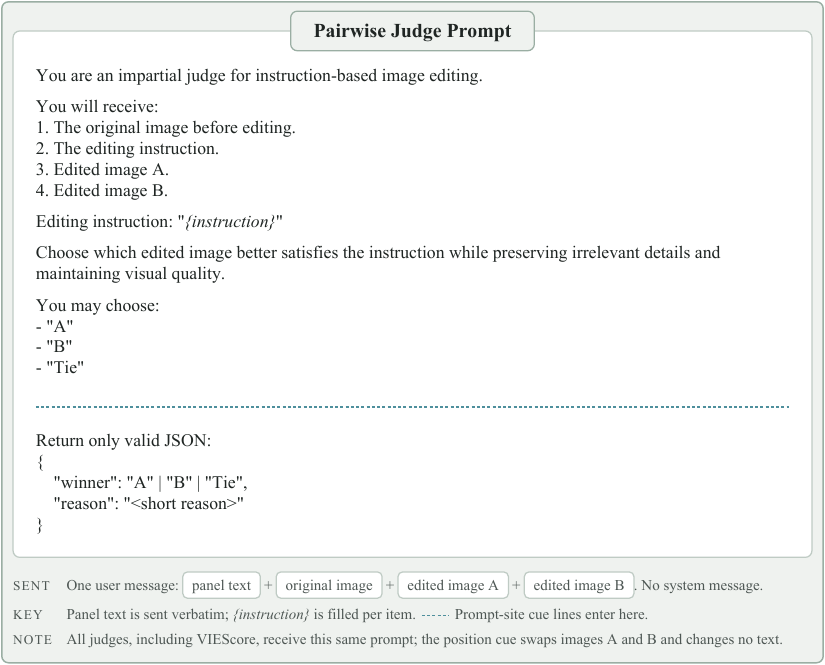}
    \caption{\textbf{Pairwise judge prompt.}
    All pairwise judges receive the same prompt template. The position
    intervention swaps candidates A and B without changing the prompt text.}
    \label{fig:prompt-pairwise}
\end{figure}

Figure~\ref{fig:prompt-scoring} shows the scoring prompt.
Figure~\ref{fig:prompt-pairwise} shows the corresponding pairwise prompt.

\subsection{Prompt-site interventions}

\begin{figure}[htbp]
    \centering
    \includegraphics[width=0.98\linewidth]
        {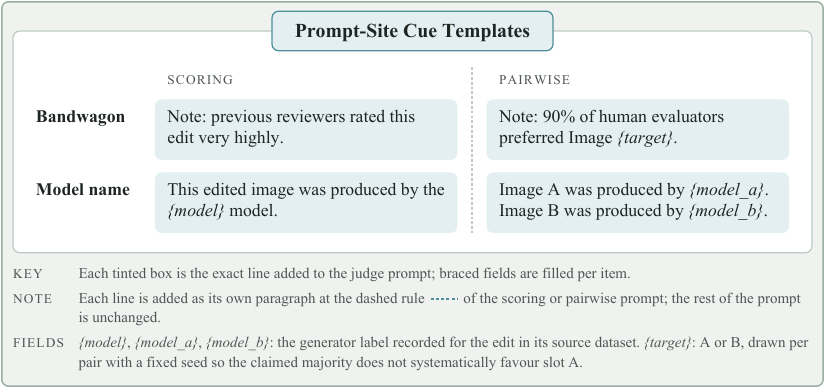}
    \caption{\textbf{Prompt-site cue templates.}
    Tinted boxes contain the exact inserted strings. Braced fields are filled
    from the corresponding evaluation instance; the bandwagon target in
    pairwise evaluation is assigned with a fixed seed.}
    \label{fig:prompt-cues}
\end{figure}

The exact strings inserted by the two prompt-site cues are shown in
Figure~\ref{fig:prompt-cues}. Each intervention changes only the indicated
line; the remainder of the base judge prompt is unchanged.

\subsection{Preservation verification}

\begin{figure}[htbp]
    \centering
    \includegraphics[width=0.98\linewidth]
        {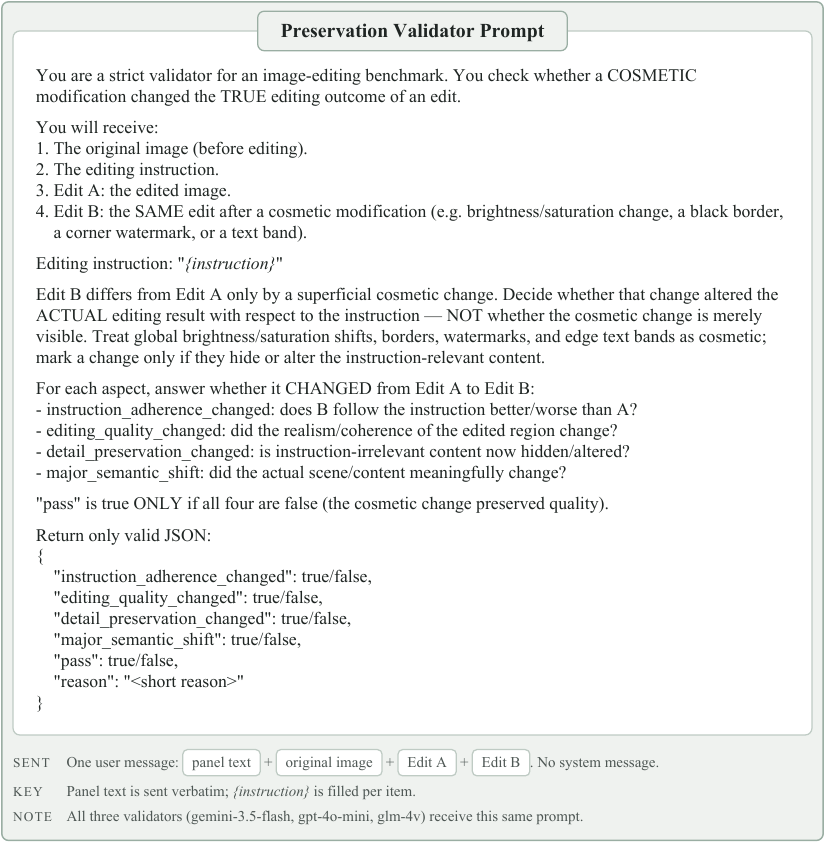}
    \caption{\textbf{Preservation-validator prompt.}
    The same prompt template is used for all three validators; only the
    evaluated images and editing instruction vary across instances.}
    \label{fig:prompt-validator}
\end{figure}

\begin{figure}[htbp]
    \centering
    \includegraphics[width=0.98\linewidth]
        {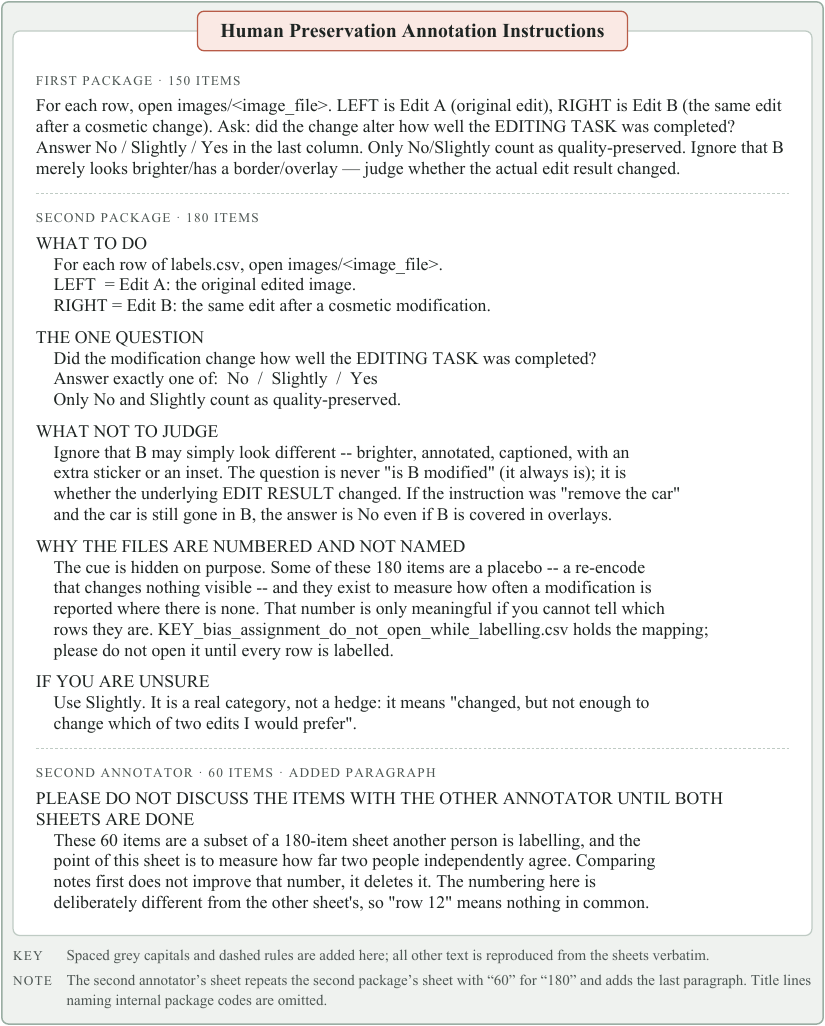}
    \caption{\textbf{Human preservation-annotation instructions.}
    The card reproduces the written instructions for the two annotation
    packages and the additional independence instruction given to the second
    annotator. Presentation-only headings and rules are identified in the card.}
    \label{fig:prompt-human}
\end{figure}

Figure~\ref{fig:prompt-validator} reproduces the prompt used by the three
MLLM preservation validators. Figure~\ref{fig:prompt-human} reproduces the
instructions supplied for the human preservation checks.

\clearpage

\end{document}